\documentclass{article}
\usepackage[utf8]{inputenc}
\usepackage{setspace}
\usepackage[margin=1in]{geometry}
\usepackage[fontsize=12pt]{scrextend}
\usepackage[T1]{fontenc}    
\usepackage{hyperref}       
\usepackage{xr-hyper}
\usepackage{url}            
\usepackage{booktabs}       
\usepackage{amsfonts}       
\usepackage{nicefrac}       
\usepackage{microtype}      
\usepackage{lipsum}		
\usepackage{graphicx}
\usepackage{natbib}
\usepackage{doi}
\usepackage{amsmath,amssymb}
\usepackage[usenames]{xcolor}
\usepackage{subcaption, multirow}
\usepackage{authblk}
\usepackage{algorithm}
\usepackage{algpseudocode}
\usepackage{xr} 
\usepackage{makecell}
\newtheorem{Proposition}{Proposition}
\newtheorem{Lemma}{Lemma}

\newtheorem{Theorem}{Theorem}

\newtheorem{Remark}{Remark}

\newtheorem{assumptionA}{Assumption}

\newcommand{\GEO}{{\text{GEO}}}

\newcommand{\rmp}{{\mathrm{P}}}
\newcommand{\I}{{\mathbf{I}}}

\newcommand{\B}{{\mathbf{B}}}
\newcommand{\Z}{{\mathbf{Z}}}

\newcommand{\bLambda}{{\boldsymbol{\Lambda}}}

\newcommand{\R}{\mathbb{R}}

\newcommand{\Q}{\mathbf{Q}}

\newcommand{\U}{{\mathbf{U}}}
\newcommand{\mH}{{\mathcal{H}}}

\newcommand{\mO}{{\mathcal{O}}}

\newcommand{\X}{{\mathbf{X}}}
\newcommand{\z}{{\mathbf{z}}}
\newcommand{\x}{{\mathbf{x}}}
\newcommand{\y}{{\mathbf{y}}}
\newcommand{\f}{{\boldsymbol{f}}}

\newcommand{\FF}{{\mathcal{F}}}

\newcommand{\bSigma}{{\boldsymbol{\Sigma}}}
\newcommand{\bXi}{{\boldsymbol{\Xi}}}

\newcommand{\supzero}{^{(0)}}
\newcommand{\supone}{^{(1)}}
\newcommand{\supa}{^{(a)}}

\def\argmindum{\mathop{\mbox{arg min}}}

\def\argmaxdum{\mathop{\mbox{arg max}}}
\def\argmax#1{\argmaxdum_{#1}}

\def\bgamma{\boldsymbol \gamma}
\def\balpha{\boldsymbol \alpha}

\def\bbeta{\boldsymbol \beta}

\def\EE{\mathbb{E}}

\def\bSig\mathbf{\Sigma}

\title{Maximin Learning of Individualized Treatment Effect on Multi-Domain Outcomes}

\author[1]{Yuying Lu$^*$}
\author[1]{Wenbo Fei\footnote{Lu and Fei contributed equally to this work.}} 
\author[1]{Yuanjia Wang\footnote{Correspondence: yw2016@cumc.columbia.edu}}
\author[2,3]{Molei Liu\footnote{Correspondence: moleiliu@bjmu.edu.cn}} 

\affil[1]{Department of Biostatistics, Columbia Mailman School of Public Health.}
\affil[2]{Department of Biostatistics, Peking University Health Science Center.}
\affil[3]{Beijing International Center for Mathematical Research, Peking University.}

\date{}
\begin{document}
\maketitle
\setstretch{1.3}
\begin{abstract}
\noindent Precision mental health requires treatment decisions that account for heterogeneous symptoms reflecting multiple clinical domains. However, existing methods for estimating individualized treatment effects (ITE) rely on a single summary outcome or a specific set of observed symptoms or measures, which are sensitive to symptom selection and limit generalizability to unmeasured yet clinically relevant domains. We propose DRIFT, a new maximin framework for estimating robust ITEs from high-dimensional item-level data by leveraging latent factor representations and adversarial learning. DRIFT learns latent constructs via generalized factor analysis, then constructs an anchored on-target uncertainty set that extrapolates beyond the observed measures to approximate the broader hyper-population of potential outcomes. By optimizing worst-case performance over this uncertainty set, DRIFT yields ITEs that are robust to underrepresented or unmeasured domains. We further show that DRIFT is invariant to admissible reparameterizations of the latent factors and admits a closed-form maximin solution, with theoretical guarantees for identification and convergence. In analyses of a randomized controlled trial for major depressive disorder (EMBARC), DRIFT demonstrates superior performance and improved generalizability to external multi-domain outcomes, including side effects and self-reported symptoms not used during training.
\end{abstract}

\setstretch{1.8}

\section{\label{sec:1} Introduction}

\subsection{Background}

Mental disorders pose challenges for precision medicine due to symptom heterogeneity and variability in patient responses. Unlike physical illnesses, often defined by clear biomarkers and objective measurements, the diagnosis of mental disorders relies on a broad spectrum of subjective symptoms, encompassing mental, cognitive, and somatic factors \citep{insel2015brain}. For instance, Major Depressive Disorder (MDD) is diagnosed based on nine core symptoms, which can manifest in over 200 distinct combinations. These diverse symptom profiles, often complicated by comorbidities, make the development of precise therapeutic strategies challenging and contribute to the low success rate of first-line treatments, with less than 50\% of patients achieving remission \citep{gartlehner2011comparative,mcgrath2020comorbidity}.

One barrier to improving treatment outcomes is the standard practice of relying on a single clinical summary to guide treatment decision-making. In mental health, treatment effects often vary across different symptom dimensions, sometimes presenting opposite effects on distinct domains \citep{maj2020clinical}. For example, Selective Serotonin Reuptake Inhibitors (SSRIs) effectively alleviate core depressive symptoms but are also associated with side effects such as sleep disturbances \citep{ferguson2001ssri}. Similarly, while antipsychotic medications may reduce hallucinations in schizophrenia, they frequently cause metabolic side effects, including significant weight gain and increased diabetes risk \citep{ballard2020role}. 
Antipsychotic medications, which may reduce hallucinations and delusions in schizophrenia, frequently cause metabolic side effects, including significant weight gain and increased diabetes risk \citep{tschoner2007metabolic,marino2015raise}. 
Relying on a single summary outcome for treatment decision-making may improve one symptom domain but harm another. Thus, it is essential to obtain reliable, generalizable estimates of the treatment effect across multiple outcome domains and latent factors to enable individualized, effective care in mental health.

\subsection{Related Literature}

Statistical methodology for estimating Individualized Treatment Effects (ITE) has advanced significantly with the development of machine learning approaches such as causal trees \citep{athey2016recursive}, causal forests \citep{wager2018estimation}, and meta-learners \citep{kunzel2019metalearners, nie2021quasi, curth2021nonparametric, kennedy2023towards}. However, these methods predominantly focus on a single scalar outcome, which is insufficient for the multifaceted nature of mental disorders. 

While multivariate regression can incorporate multiple observed outcomes,
high-dimensional and subjective symptom data in psychiatry are more effectively modeled using latent variable models and factor analysis \citep{moustaki2000generalized, muthen2002beyond}. In this context, Item Response Theory (IRT) and generalized factor models are widely used to identify underlying constructs from categorical symptoms \citep{embretsonreise, reise2009item}, with recent advances extending these to high-dimensional settings via joint maximum likelihood estimation \citep{chen2019joint,zhang2020note}.

To inform treatment decision-making with multi-domain outcomes, existing approaches often aggregate sub-domain outcomes into a single global measure using fixed weights. For instance, recent works have proposed preference-sensitive aggregation \citep{butler2018incorporating,luckett2021estimation} or composite outcomes \citep{chen2020representation,chen2021learning} to learn individualized treatment rules. Also, \citet{nwankwo2024reduced} proposed a reduced-rank multi-objective policy learning method leveraging low-dimensional representations of the outcomes. However, these approaches assume a fixed aggregation function, thereby ignoring unobserved or infrequently observed outcomes that could yield a ``worst-case'' ITE. This is a critical limitation in risk-sensitive clinical settings, calling for more generalizable methods \citep{cinelli2025}.

This challenge relates to the growing literature on maximin effects and distributionally robust learning (DROL), which seeks to optimize performance under the worst-case shift in data distribution \citep[e.g.]{meinshausen2015maximin,sagawa2019distributionally,wang2023distributionally}. While such adversarial learning frameworks have been applied to ITE and policy learning to address heterogeneity across data sources \citep{shi2018maximin,2024Minimaxi}, they have not been adapted to balance trade-offs across latent symptom domains. These methods optimize over the convex hull of multiple observed outcomes, which is often overly conservative and, more importantly, problematic if the measured symptoms are a ``biased'' subset of the full symptom spectrum. By restricting optimization to observed items, this strategy fails to generalize to unmeasured but relevant domains that exist within the full ``hyper-population'' of symptoms (see Section \ref{sec:method:geo}).

\subsection{Our Contribution}

To address the limitations of using a single outcome for treatment optimization in mental health, we propose the multi-Domain Robust estimation of Individualized and Factorized Treatment effect (DRIFT) framework. This approach leverages generalized factor analysis to identify latent disease constructs from multi-dimensional symptom data and constructs an ``on-target'' set of clinically plausible outcomes anchored by a Global Evaluation Outcome (GEO). By optimizing the worst-case predictive performance across this set, DRIFT ensures that the estimated treatment effects are stable and balanced across diverse symptom domains. Our main contributions can be summarized as follows.

{\bf Novel Framework:} 
We propose a maximin learning framework that extrapolates the GEO-anchored uncertainty set in latent space to account for unmeasured or underrepresented clinical symptoms. Unlike standard group maximin regression methods deterministically restricted to the empirical convex hull of observed symptom outcomes, this GEO-guided extrapolation renders the solution distributionally robust to variability in observed symptom coverage. It also yields a more generalizable ITE estimate by recovering the full spectrum of potential symptoms, thereby protecting DRIFT against failures within clinical domains underrepresented in the observed data.

{\bf Convenient Estimation Procedures with Theoretical Guarantees:} We provide a computationally efficient, three-step algorithm for identifying the optimal treatment effect estimator. We establish that our maximin problem admits a closed-form solution and prove the consistency of the resulting estimator. Crucially, we show that our estimator is not affected by the inherent non-identifiability of latent factor models, remaining invariant to their rotational indeterminacy.

{\bf Enhanced Clinical Generalizability in Real World:} We demonstrate the practical utility of DRIFT through an analysis of the EMBARC \citep{trivedi2016establishing} randomized controlled trial for Major Depressive Disorder (MDD). Our results show that DRIFT achieves superior generalizability to external multi-domain outcomes (e.g., side effects and self-reported symptoms) not used during model training, highlighting its potential to prevent the neglect of critical clinical factors.

\subsection{Outline of the Paper}

We introduce the key concepts and set up the DRIFT framework in Section \ref{s:setup}. Based on this, we develop an estimation method for our proposal in Section \ref{sec:est:method} and justify its theoretical properties in Section \ref{sec:theory}. In Section \ref{sec:simul}, we conduct extensive simulation studies to evaluate the finite-sample performance of DRIFT and compare it with existing approaches. In Section \ref{sec:real}, we apply DRIFT to analyze the real-world data collected from a clinical trial studying the treatment effect heterogeneity of Sertraline on major depressive disorder (MDD). In Section \ref{s:discuss}, we conclude the paper with a discussion. Technical proofs of the theoretical results and additional numeric results are provided in the Supplementary Material.

\section{Conceptual Framework}
\label{s:setup}

\subsection{Problem formulation}

Let $\mathbf{X} = (X_1, \cdots, X_p)$ denote baseline covariates and treatment effect moderators of the outcomes to be introduced next. Suppose $K$ latent factors underlie a subject's overall mental health status. Let $\mathbf{Z}^{(a)} = (Z_1^{(a)}, \ldots, Z_K^{(a)})$ denote the potential outcomes of these latent factors after receiving treatment $A = a\in \{0,1\}$, where $Z_k^{(a)}\in \mathbb{R}$, $k = 1, \ldots, K$ represents the $k$th latent variable. Let $\mathbf{Y}^{(a)} = (Y_1^{(a)}, \ldots, Y_J^{(a)})$ denote the potential outcomes of the observed multi-domain symptom measures after receiving treatment $a$, where $Y_j^{(a)}$ represents the $j$th symptom and it can be discrete or continuous. While the latent factors are not explicitly observed, each measured symptom, 
e.g., item responses from the Hamilton Rating Scale for Depression (HAM-D), is a partial manifestation of the subject's underlying latent states, e.g., depression severity, general anxiety level, and psychotic propensity. Typically, the total number of observed outcomes $J$ is much larger than the number of latent factors $K$.

We assume that (C.1) {\em measurement conditional independence}: $\mathbf{Y}^{(a)}$ is independent of $(A,\X)$ conditional on $\mathbf{Z}^{(a)}$; and 
(C.2) {\em measurement invariance}: the conditional distributions of $\mathbf{Y}^{(1)}\mid\mathbf{Z}^{(1)}$ and $\mathbf{Y}^{(0)}\mid\mathbf{Z}^{(0)}$ are the same. (C.1) implies that baseline covariates $\mathbf{X}$ and treatment $A$ affect the observed outcomes $\mathbf{Y}$ only through the latent factors in $\mathbf{Z}$. 
(C.1) and (C.2) are commonly used on item response data from clinical studies of mental health \citep{harman1976modern,  lohr2002assessing, chen2021learning}. Additionally, we assume 
(C.3) {\em randomized treatment assignment}: $A\perp(\X,\mathbf{Z}^{(0)},\mathbf{Z}^{(1)},\mathbf{Y}^{(0)},\mathbf{Y}^{(1)})$; 
and (C.4) {\em stable unit treatment value assumption (SUTVA)}: $\mathbf{Y} = \mathbf{Y}^{(a)}$ and 
$\mathbf{Z} = \mathbf{Z}^{(a)}$ when $A=a$. (C.3) could be replaced by the weaker no unmeasured confounding assumption $({\bf Z}^{(a)}, {\bf Y}^{(a)})\perp A\mid {\bf X}$ in observational studies (see Section \ref{sec:method:obs}). We introduce the following generalized factor model where $Y^{(a)}_1, \ldots, Y_J^{(a)}$ are conditionally independent given $\mathbf{Z}^{(a)}$ and satisfy that for $a\in\{0,1\}$,
\begin{align}
    \mathbb{E}[Y^{(a)}_{j} | \mathbf{Z}^{(a)}=\mathbf{z}] =g^{-1}_j(\mathbf{z}^{\top}\boldsymbol{\alpha}_j-\zeta_j).
    \label{eq:factor}
\end{align}
Here, each $Y^{(a)}_{j}$ is associated with a linear function of the latent factors $\mathbf{Z}^{(a)}$ through the coefficient $\boldsymbol{\alpha}_j{\in \R^K}$, the intercept $\zeta_j \in \R$ and some known link function $g_j(\cdot)$ that monotonically increases. Also, assume the potential outcomes of the latent factors $\mathbf{Z}^{(a)}$ follow that:
\begin{align}
     \mathbf{Z}^{(a)} =\boldsymbol{f}^{(a)}(\mathbf{X})+\boldsymbol{\epsilon}^{(a)},
    \label{eq:Z}
\end{align}
where the vector function $\boldsymbol{f}^{(a)} = (f_1^{(a)}, \ldots, f_K^{(a)})$ encodes the treatment response models of $\mathbf{Z}^{(a)}$, and $\boldsymbol{\epsilon}^{(a)}$ are random components being mean-zero and independent of $\mathbf{X}$. 

\subsection{Factorized outcomes and ITE}
\label{sec:factorized_outcome}

We define the class of latent factor representations as $\mH:= \{\z\mapsto\boldsymbol{\gamma}^\top\mathbf{z}-\zeta\mid \bgamma\in \R^K, \zeta\in\R\}$, which contains all linear functions of the $K$-dimensional latent factors. For any observed clinical symptom $Y$, we define its best latent representation in $\mH$, denoted by $\phi_{Y}(\z)$, as the minimizer of the population expected loss:
\begin{equation}
    \phi_{Y}(\z):= \bgamma_Y^\top\z-\zeta_Y=\argmindum_{\phi\in\mH}\ \EE\big[\ell_Y(Y,\phi(\Z))\big].\notag
\end{equation} 
The loss function $\ell_Y(y,u)$ is assumed to be convex with respect to $u$ and specified based on the type of outcomes: for continuous $Y$, we use the squared error $\ell_Y(y,u) = (y-u)^2$; for binary $Y$, we use the negative log-likelihood $\ell_Y(y, u) = -yu + \log(1 + \exp(u))$. We further provide the choice of the link and loss functions for ordinal item responses with multiple categories in Appendix. Under assumptions (C.1) through (C.4) and the factor model~\eqref{eq:factor} for each observed item $Y_j$, these choices ensure that $\phi_{Y_j}(\z)= g_j(\EE[Y_j\mid \Z=\z])=\mathbf{z}^{\top}\boldsymbol{\alpha}_j-\zeta_j$, with the link $g_j(u)=u$ for continuous $Y_j$ and $g_j(u)=\log\{u/(1-u)\}$ for binary $Y_j$.

To quantify the deviation of any candidate linear representation $\phi(\z)$ from a specific outcome $Y$, we define the excess loss as
\begin{equation}
d_Y(\phi)=\EE\left[\ell_Y(Y,\phi(\Z))\right]-\EE\left[\ell_Y(Y,\phi_Y(\Z))\right].
 \label{eq:d_Y}
\end{equation}
This non-negative metric serves as a distance measure and will be used to construct a set of clinically plausible outcomes to be defined later. For binary $Y$, $d_Y(\phi)$ is equivalent to the expectation of the Kullback–Leibler (KL) divergence between the Bernoulli distributions with success probabilities
${\rm expit}(\phi_Y(\Z))$ and ${\rm expit}(\phi(\Z))$ where ${\rm expit}(u)=e^u/(1+e^u)$.

In application fields like psychiatry and psychology, the ITE on any factorized outcome $\phi(\Z)$ for a subject with baseline covariates $\x$ is typically defined as:
\begin{align}
     \tau_{\phi}(\mathbf{x}):&=\EE\left[\phi(\mathbf{Z}^{(1)})-\phi(\mathbf{Z}^{(0)})\mid\mathbf{X}=\mathbf{x}\right] =\boldsymbol{\gamma}^\top\Big(\boldsymbol{f}^{(1)}(\mathbf{x})-\boldsymbol{f}^{(0)}(\mathbf{x})\Big),
     \label{eq_CATE}
\end{align}
with $\boldsymbol{f}\supa(\x)=\EE\big[\Z\supa\mid\mathbf{X}=\mathbf{x}\big]$. In this work, we consider a linear model for the treatment effect $\f^{(a)}(\X) = \bLambda^{(a)}\X, \ \bLambda^{(a)}\in\R^{K\times p}$. The framework readily extends to nonlinear models.

To ensure that the weights in model \eqref{eq_CATE} are clinically interpretable, we leverage a one-dimensional Global Evaluation Outcome (GEO), denoted by $O$, that captures the overall effectiveness of an intervention. For example, the Clinical Global Impression (CGI) Scale is routinely collected in mental health studies \citep{busner2007clinical}, used as the GEO to assess overall improvement of a patient. When $O$ is observed or well-defined by domain knowledge, we define the factorized GEO representation as:
\begin{equation}
\phi_\GEO(\z):=\boldsymbol{\gamma}_{\text{GEO}}^\top\mathbf{z}-\zeta_{\text{GEO}}=\argmindum_{\phi\in\mH}\ \EE\left[\ell_O\big(O,\phi(\Z)\big)\right],
\label{equ:geo}
\end{equation}
and the corresponding ITE function as $\tau_{\phi_{\text{GEO}}}(\x)=\EE\big[\phi_\GEO(\mathbf{Z}^{(1)})-\phi_\GEO(\mathbf{Z}^{(0)})\mid\mathbf{X}=\x\big]
$. In the absence of such a GEO, we propose a minimax strategy to obtain $\phi_{\text{GEO}}$ as the ``center'' of the observed responses $Y_j$'s (see Section~\ref{sec:on-target}).

\subsection{On-target set of factorized outcomes}\label{sec:frame:ontar}

\begin{figure*}[htb!]
  \centering
  \begin{subfigure}[t]{0.5\textwidth}
     \centering
     \includegraphics[width=\linewidth]{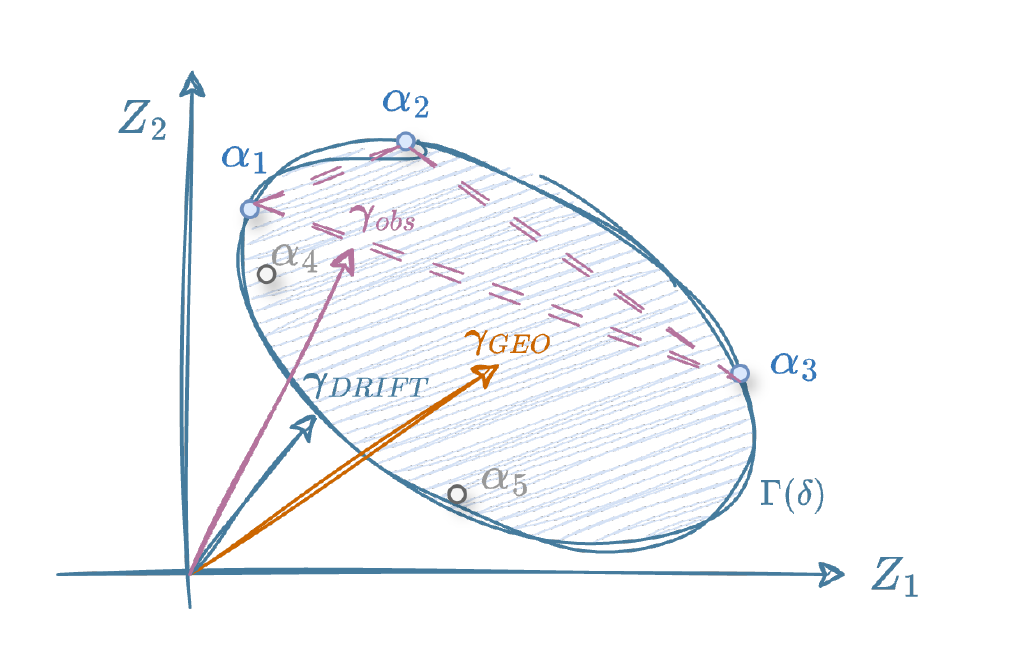}
     \caption{Geometry of aggregation weights in the latent space ($Z_1, Z_2$). $\{\balpha_i\}_{i=1}^3$ (observed), $\{\balpha_i\}_{i=4}^5$ (unobserved) loading vectors representing the specific multi-domain outcomes. The pink quadrangle is the convex hull of the observed outcomes, over which standard maximin methods optimize. It yields the ``observed-only'' solution $\bgamma_{\text{obs}}$. $\bgamma_{\text{GEO}}$ (orange arrow)is the GEO aggregation weight, serving as the clinical anchor. $\Gamma(\delta)$ (shaded ellipses) denotes the robust ``on-target'' uncertainty sets with radius $\delta$, encompassing weights close to $\bgamma_{\text{GEO}}$. $\bgamma_{\text{DRIFT}}$ denotes the optimal weight of DRIFT estimator, which balances performance over $\Gamma(\delta)$ rather than only the observed hull.}
     \label{fig1:maxmin}
  \end{subfigure}\hspace{1em}
  \begin{subfigure}[t]{0.42\textwidth}
     \centering
     \includegraphics[width=\linewidth]{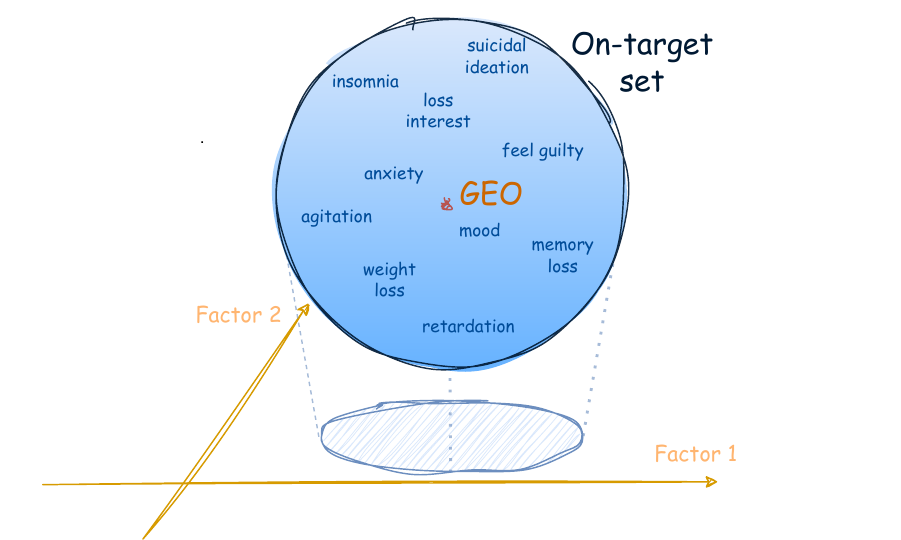}
     \caption{Clinical Interpretation of the On-Target Set: Mapping of the geometric uncertainty set to clinical domains. GEO (Center): The global improvement measure (e.g., CGI). On-target set of outcomes (Blue Circle): The set of clinically plausible outcomes extrapolated from the GEO. Unlike the sparse observed data, this set captures unmeasured but relevant domains residing in the GEO's latent neighborhood, ensuring the treatment effect is robust to these overlooked factors.}
     \label{fig1:GEO}
  \end{subfigure} 
  \caption{Illustration of the DRIFT Framework and Concepts.}
\end{figure*}

In precision mental health, accounting for symptoms and outcomes across multiple domains is crucial to guide treatment decisions. Overlooking an important domain can lead to suboptimal or ineffective decisions. Although the GEO-oriented factorized representation $\phi_{\text{GEO}}$ and ITE function $\tau_{\boldsymbol{\phi}_{\text{GEO}}}$ are clinically informative, they remain tied to specific observed outcomes, thus lack generalizability to capture the full spectrum of multi-domain outcomes in mental disorders. Therefore, it is essential to maintain a robust and balanced characterization of the ITE across a diverse array of outcomes represented by latent factors $\mathbf{Z}$. For this purpose, we introduce a key concept called the {\em on-target} set of representation: 
\begin{align}
\label{ITE:eq:set}
\Phi(\delta) = \left\{\phi\mid d_O(\phi)\le \delta,~\phi\in\mH\right\}, 
\end{align}
where $d_O(\cdot)$ denotes the excess loss for the GEO response as defined in \eqref{eq:d_Y} (replacing $Y$ with $O$), and $\delta\ge0$ is a radius parameter controlling the allowable deviation in representing GEO among the candidate representations included in $\Phi(\delta)$. Each factorized outcome $\phi(\mathbf{Z})$, with $\phi \in \Phi(\delta)$, constitutes a clinically meaningful aggregation of the underlying latent factors, anchored by $\phi_\GEO(\Z)$ yet distinct from it. A small $\delta$ restricts consideration to outcomes very similar to $\phi_\GEO(\Z)$ while a larger $\delta$ allows a broader range of factorized outcomes. Correspondingly, we define the set of aggregation weights associated with $\phi\in \Phi(\delta)$ as:
\begin{align}
\label{eq:Gamma_set}
\Gamma(\delta) = \left\{\bgamma \ \Big| \  \min_{\zeta\in \R}
\{d_O(\phi)\mid \phi:\z\mapsto \bgamma^\top\z-\zeta\}\le \delta, ~\bgamma\in \R^K\right\}.
\end{align}
When $\delta = 0$, the uncertainty set collapses to a single clinical anchor (GEO), simplifying the approach to a classical estimator that considers a fixed weight $\boldsymbol{\gamma}_{\text{{GEO}}}$ (Figure~\ref{fig1:maxmin}). As $\delta$ increases, the set expands to a latent neighborhood surrounding the GEO. Clinically, this expanded ``on-target'' set captures a comprehensive spectrum of relevant outcomes that may be unmeasured or underrepresented in the observed $Y_1,\ldots,Y_J$, such as individual symptoms reported in HAM-D or burden of side effects (Figure~\ref{fig1:GEO}). {Further geometric interpretations of GEO-anchored extrapolation are provided in Section \ref{sec:method:geo} and our real-world study.} 

\subsection{Maximin framework}

We now introduce a novel maximin framework for estimating ITE that achieves stable, balanced predictive performance across a range of symptom representations in $\Phi(\delta)$. This enables a generalizable evaluation of treatment effects and more reliable treatment recommendations. For any factorized representation $\phi( \z)= \boldsymbol{\gamma}^\top\mathbf{z}-\zeta$ and a learner $\tau(\mathbf{x})$ belonging to some prespecified model space $\mathcal{F}$, we introduce the reward function
\begin{align*}
{R}_{\phi}\left(\tau\right):
=&\EE\Big[\big(\phi(\Z\supone)-\phi(\Z\supzero)\big)^2-\big(\phi(\Z\supone)-\phi(\Z\supzero)- \tau(\mathbf{X})\big)^2\Big]\\
=&\EE\Big[\big(\tau_{\phi}(\mathbf{X})\big)^2-\big(\tau_{\phi}(\mathbf{X})- \tau(\mathbf{X})\big)^2\Big].   
\end{align*}
This quantity measures how much of the variation in the ITE function $\tau_{\phi}(\mathbf{x})$ (defined in \eqref{eq_CATE}) is captured by the candidate estimator $\tau(\mathbf{x})$. A higher value of ${R}_{\phi}\left(\tau\right)$ indicates that $\tau(\mathbf{x})$ better predicts ITE on the factorized outcome $\phi(\Z)$. To obtain a robust characterization of the ITE over all on-target factorized representations in $\Phi(\delta)$ defined in \eqref{ITE:eq:set}, 
we propose a maximin optimization to obtain the estimator that maximizes the worst-case reward ${R}_{\phi}\left(\tau\right)$:
\begin{align}\label{eq:DRIFT}
\tau^*:=\argmax{\tau\in\mathcal{F}}\min_{\phi\in\Phi(\delta)} {R}_{\phi}\left(\tau\right),
\end{align}
which we call the multi-Domain Robust estimator of the Individualized and Factorized Treatment effect (DRIFT). This formulation is an adversarial learning problem that pursues a balance (Nash equilibrium) between the ITE function $\tau$ and the representation $\phi$. 
At this equilibrium, the solution $\tau^*$ achieves a stable prediction performance over all representations in $\Phi(\delta)$ in the sense that it maximizes the worst-case ITE prediction performance. We will also demonstrate in Proposition \ref{propo:ITR} that optimizing the worst-case reward ${R}_{\phi}\left(\tau\right)$ leads to the optimal individualized treatment regime for the outcome $\phi(\Z)$ under certain conditions.

Leveraging Sion's Minimax Theorem \citep{sion}, Theorem \ref{ITE:thm1} establishes that problem (\ref{eq:DRIFT}) has an explicit-form solution formed as a weighted combination of the latent-factor-wise ITE functions $\big\{\tau_k:k\in[K]\big\}:= \big\{f_k^{(1)}-f_k^{(0)}:k\in[K]\big\}$.
\begin{Theorem}[Identification] Suppose the function class $\mathcal{F}$ is convex with $\tau_k\in \mathcal{F}$ for $k\in[K]$, then $\tau^* $ defined in (\ref{eq:DRIFT}) is identified as:
\label{ITE:thm1}
\begin{align}
\label{eq:quad}
    \tau^*(\mathbf{x}) = \sum_{k=1}^K \gamma^*_k\tau_k(\mathbf{x}) \quad \text{with} \quad \boldsymbol{\gamma}^*= \argmindum_{\boldsymbol{\gamma}\in\Gamma(\delta)}\ \boldsymbol{\gamma}^\top\bXi \boldsymbol{\gamma},
\end{align}
where $\Xi_{k,l}=\mathbb{E}[\tau_k(\mathbf{X})\tau_l(\mathbf{X})]$ for $k,l\in [K]$.  
\end{Theorem}
Theorem \ref{ITE:thm1} implies a straightforward way to compute the maximin ITE function $\tau^*$. In particular, when the ITE function $\tau_k$ for each latent factor $Z_k$ is available, one only needs to solve the quadratic programming problem in (\ref{eq:quad}) to obtain the optimal adversarial weights $\boldsymbol{\gamma}^*$. The final estimator is then formed by linearly aggregating $\tau_k$'s. This result shares a similar spirit with the group maximin regression \citep{meinshausen2015maximin} aiming to combine models from multiple sources through adversarial learning, with its solution being some convex combination of the source-specific models. 

Interestingly, the strategy of linearly aggregating informative latent factors has been widely used to address multi-domain outcomes in areas such as mental health \citep{qiu2018estimation}. An advantage of such linear ITE functions is that they provide a straightforward interpretation, with the linear weights quantifying the relative contributions of the latent factors to the aggregated model. Also, it facilitates easy implementation of decision rules in clinical practice. 

\begin{Remark}
In our framework, it is essential to define the on-target set $\Phi(\delta)$ in (\ref{eq:DRIFT}) using the low-dimensional representation $\mathbf{Z}$ instead of applying linear combinations directly to the original outcomes $\boldsymbol{Y}$. First, the original outcomes may have different distributional types and scales, whereas {the GEO-guided on-target factorized outcome subspace $\{\phi(\Z)\mid \phi\in \Phi(\delta)\}$} is free of these issues because it is invariant to any scaling or rotation of $\mathbf{Z}$ (Proposition \ref{propo:well_defined}). Second, using  $\mathbf{Z}$ effectively overcomes the high-dimensionality and measurement noise of $\boldsymbol{Y}$.    
\end{Remark}

\subsection{Geometric interpretation {of GEO-anchored extrapolation}}\label{sec:method:geo}

In clinical practice, the full spectrum of potential symptoms constitutes a ``hyper-population'' of outcomes, whereas any given study collects only a subset determined by its design. As a concrete example, in the clinical trial of major depressive disorder (MDD; Section~\ref{sec:real}), mania-related symptoms (e.g., feeling unusually euphoric or energized, talking more than usual, having a flood of exciting ideas) are not represented in the Hamilton Depression Rating Scale (HAM-D), a primary outcome in many MDD studies. {As a result, ITE estimators based only on observed outcomes may overlook underrepresented domains and be sensitive to the particular symptoms included in the study. 
Interestingly, DRIFT overcomes these limitations through the GEO-anchored extrapolation strategy. The geometric interpretation involves two main aspects:} 

\textbf{The GEO-anchored set $\Gamma(\delta)$ reduces algorithmic sensitivity:} 
{Group maximin methods \citep{meinshausen2015maximin} were originally developed for regression on a common outcome across multiple data sources.} In analogy to this strategy, under our setting, the ``observed-only'' maximin approach (\textit{Obs Maximin}) constrains its uncertainty set to the convex hull of aggregation weights for all the observed factorized outcomes (pink triangle in Figure~\ref{fig1:maxmin}). It yields an estimator $\gamma_{\text{obs}}$ that could be highly sensitive to a subset of outcomes from the hyper-population measured in some specific study. In contrast, \textit{DRIFT} optimizes over a GEO-anchored robust region $\Gamma(\delta)$ that recovers the ``hyper-population'' of outcomes. As illustrated in Figure~\ref{fig1:maxmin}, introducing unobserved weights (e.g., $\alpha_4$ and $\alpha_5$) can significantly reshape the convex hull of the observed outcomes' representation while only mildly perturbing $\Gamma(\delta)$, yielding a more stable solution $\bgamma_{\text{DRIFT}}$.

\textbf{Extrapolation enhances generalizability:} Confined to the convex hull of observed outcomes, the observed-only method generalizes poorly to underrepresented or unmeasured clinical outcomes. For instance, if observed symptoms preferentially load on latent factor $Z_2$ over $Z_1$ (Figure~\ref{fig1:maxmin}), the resulting ITE estimators overlook treatment effects operating through $Z_1$. Consequently, it fails to detect beneficial effects on unmeasured symptoms or adverse side effects hidden within underrepresented latent dimensions. {\textit{DRIFT} mitigates these failures by extrapolating to the larger and more comprehensive ``hyper-population'', thereby accounting for plausible outcomes in underrepresented domains. This yields a more generalizable characterization of treatment effects.} We validate this advantage empirically in the HAM-D study (Section~\ref{sec:real}), where the observed-only method suffers from failures in the underrepresented mania domain, whereas \textit{DRIFT} maintains robustness.

We also note that defining the uncertainty set via the excess loss $d_O$ ensures that our ITE estimator is invariant to the rotational indeterminacy of the latent factors $\Z$. The uncertainty set of the \textit{Obs Maximin} method is defined strictly on the coordinates of estimated loadings of the observed items, which may require more assumptions, e.g. the varimax criterion \citep{kaiser1958varimax}, to ensure proper interpretation and identification. Free from this issue, DRIFT remains well-defined even when the ``true'' underlying $\Z$ is not identifiable (see Proposition \ref{propo:well_defined}).

\section{Estimation Method}\label{sec:est:method}

\subsection{Generalized factor analysis}
\label{sec:fct_model}
Suppose that for subject $i\in\{ 1,\dots, N\}$, we observe baseline covariates ${\x_i} = (x_{i1},\dots,x_{ip})\in \R^{p}$, received treatment $a_i\in\{0,1\}$ and multi-domain outcomes $\y_i = (y_{i1},\dots,y_{iJ})\in \R^{J}$. Denote the unobserved latent factors as $\z_i=(z_{i1},\dots, z_{iK})\in \R^K$. In (\ref{eq:JML}), we introduce the constrained joint maximum likelihood estimation (CJMLE) approach \citep{chen2019joint}:
\begin{align}  
\big\{\widehat{\z}_i,\widehat{\balpha}_j,\widehat{\zeta}_j: i \in[N], j\in[J]\big\}=&\  \argmindum_{\z_i,{\balpha}_j,{\zeta}_j} \sum_{j=1}^J \sum_{i=1}^N \ell_{Y_j}(y_{ij},\balpha_j^\top\z_i-{\zeta}_j), \notag \\  \quad s.t.\quad {1+\|\z_i\|^2}\le C^2,  \ \ & {\zeta_j^2+\|\balpha_j\|^2} \le C^2,\quad \forall i \in[N], j\in[J],
\label{eq:JML}
\end{align}
where $\{\ell_{Y_j}(\cdot,\cdot):j\in[J]\}$ are loss functions for the observed outcomes $Y_j$'s as defined in Section \ref{sec:factorized_outcome}  and $C>0$ is a pre-specified regularization parameter, set as $5\sqrt{K}$ in our numerical studies. In practice, the latent space dimension $K$ can be chosen based on expert knowledge or information criterion. To optimize (\ref{eq:JML}), we adopt the alternating minimization procedure described in Appendix, with the constrained optimization problem handled by projected gradient descent.

With the observed $(\x_i,a_i)$ and the latent factors $\widehat{\z}_i$ estimated by (\ref{eq:JML}), we then learn $\widehat{\boldsymbol{f}}^{(a)} =(\widehat f_1^{(a)}, \dots, \widehat f_K^{(a)})~(a\in \{0,1\})$ as estimates of the treatment response functions defined in (\ref{eq:Z}). This is a standard regression problem, where both traditional methods (e.g., linear regression) and machine learning methods (i.e., neural networks) can be used. Because the latent $\widehat{\z}_i$'s are only identifiable up to an invertible rotation and a constant shift, $\widehat{\boldsymbol{f}}^{(a)}$ (and, accordingly, $\widehat{\tau}_k = \widehat{f}_k^{(1)}-\widehat{f}_k^{(0)}$) can be consistent with $\f^{(a)}$ in (\ref{eq:Z}) (and $\tau_k = f_k^{(1)}-f_k^{(0)}$) up to corresponding reparameterizations. Consequently, the empirical estimators $\widehat{\z}_i$ and $\widehat{\tau}_k$ can be directly used to solve the downstream maximin problem (see Proposition \ref{propo:well_defined}).

\subsection{Constructing the on-target set}
\label{sec:on-target}

Construction of the on-target set $\Gamma(\delta)$ requires specifying the GEO representation $\phi_{\text{GEO}}\in \mH$ and selecting the radius parameter $\delta$. The representation $\phi_{\text{GEO}}$ serves as the ``center'' of $\Gamma(\delta)$. Let $o_i$ denote the GEO of subject $i$. When the GEO is available in the data, we solve the empirical version of (\ref{equ:geo}) to obtain $\widehat{\phi}_\GEO(\z) =\widehat{\bgamma}_\GEO^\top\z-\widehat\zeta_\GEO$ as an estimate of ${\phi}_\GEO(\z)$.

When GEO is unobserved, we propose to learn $\widehat{\phi}_{\text{GEO}}$ as the center of the observed outcomes' representations $\big\{\widehat{\phi}_{Y_j}(\z)=\widehat{\balpha}_j^\top\z-\widehat{\zeta}_j:j\in[J]\big\}$. Specifically, we solve the minimax problem:
\begin{equation}
\widehat{\phi}_\GEO=\argmindum_{\phi\in\mH}\ \max_{j\in [J]} \frac{1}{N}\sum_{i=1}^N\big(\phi(\widehat{\z}_i)-\widehat{\phi}_{Y_j}(\widehat{\z}_i)\big)^2,
\label{equ:minimax:center}
\end{equation}
which minimizes the largest distance to the factors of all observed outcomes to find their center. For this problem, one can again invoke Sion's Minimax Theorem \citep{sion} to derive an explicit-form solution (see Theorem~\ref{ITE:thm1}). Note that problem (\ref{equ:minimax:center}) coincides with the group min-max regret framework proposed by \cite{mo2024minimax}. However, an important difference is that we use the solution of (\ref{equ:minimax:center}) as an anchor, or center, for constructing the on-target uncertainty set rather than using it as the final estimator.

The radius parameter $\delta$ determines the size of the uncertainty set for the maximin problem (\ref{eq:DRIFT}). If $\delta$ is too small, the on-target set may become overly restrictive, limiting the generalizability of DRIFT to those outcomes nearly identical to the GEO. Conversely, if $\delta$ is too large, the set may include outcomes that are not clinically meaningful or relevant, leading to overly conservative solutions. To achieve a balanced trade-off, we recommend setting the radius parameter as
\(
\delta^{\dagger}=\min\Big\{\max_{j\in[J]} \widehat{d}_{O}(\widehat{\phi}_{Y_j}),~0.95\min_{\zeta\in\R}\widehat{d}_{O}(\zeta)\Big\},
\) where $$\widehat{d}_{O}(\phi):=\frac{1}{N}\sum_{i=1}^N\left[\ell_O(o_i,\phi(\widehat\z_i))-\ell_O(o_i,\widehat\phi_\GEO(\widehat\z_i))\right]$$ 
denotes the empirical excess loss operator of the GEO $O$ as defined in (\ref{eq:d_Y}). Note that even without the observation of $O$, one can still compute $\widehat{d}_{O}$ with $\widehat{\phi}_\GEO$ obtained by (\ref{equ:minimax:center}) because $O$ can be replaced by its conditional expectation specified by $\widehat{\phi}_\GEO(\Z)$ under the generalized factor model. This threshold $\delta^{\dagger}$ is selected to be large enough to encompass the observed factorized outcomes, thereby addressing the limited observed symptom coverage illustrated in Section \ref{sec:method:geo}. In addition, $\delta^{\dagger}$ is bounded by $0.95\min_{\zeta\in\R}\widehat{d}_{O}(\zeta)$ to ensure that our robust estimator does not degenerate to a null model. The on-target set for aggregation weights of the factorized outcomes is constructed as $\widehat{\Gamma}(\delta^{\dagger})$, where $\widehat{\Gamma}(\delta)$ denotes the empirical on-target set obtained from~\eqref{eq:Gamma_set} by replacing $d_O$ with its empirical estimate $\widehat{d}_O$.

\subsection{Maximin estimation}
\label{sec:method_2}

We now combine the components from the previous sections to compute the DRIFT estimator. Our goal is to empirically estimate the maximin-optimal ITE function $\tau^*$ defined in (\ref{eq:DRIFT}), which achieves robust performance on the data-driven on-target set. Following Theorem \ref{ITE:thm1}, we solve the quadratic optimization problem: 
\begin{equation}
\widehat{\boldsymbol{\gamma}} = \argmindum_{\boldsymbol{\gamma}\in\widehat{\Gamma}(\delta)}\boldsymbol{\gamma}^\top\widehat{\boldsymbol{\Xi}} \boldsymbol{\gamma},\  \text{ where } \widehat{\boldsymbol{\Xi}} = (\widehat{\Xi}_{k,l})_{K\times K},~\widehat{\Xi}_{k,l} = N^{-1} \sum_{i=1}^{N} \widehat{\tau}_k\left(\mathbf{x}_i\right) \widehat{\tau}_l\left(\mathbf{x}_i\right),
\label{equ:emp:maximin}
\end{equation}
and derive the final estimator for $\tau^*$ as $\widehat{\tau}(\mathbf{x}) = \sum_{k=1}^K \widehat{\gamma}_k \widehat{\tau}_k(\mathbf{x})$. In practice, the radius parameter $\delta$ used in (\ref{equ:emp:maximin}) is recommended as $\delta^{\dagger}$ introduced in Section~\ref{sec:on-target}. The complete procedure for obtaining the DRIFT estimator is outlined in Algorithm~\ref{algo:maximin}.

\begin{algorithm}[htb!]
{
\caption{Empirical estimation procedures for DRIFT.}
\label{algo:maximin}
\begin{algorithmic}[1]
    \State \textbf{Input}: Data $\{\mathbf{x}_i, \y_i, o_i\}_{i=1}^N$ and latent dimension $K$.
    \State \textbf{Step 1:} Solve (\ref{eq:JML}) (with $C=5\sqrt{K}$) to obtain latent factors $\{\widehat{\z}_i\}_{i=1}^N$ and ITE estimators $\{\widehat{\tau}_k\}_{k=1}^K$.
    \State \textbf{Step 2:} Obtain $\widehat{\bgamma}_\text{GEO}, \widehat{\zeta}_\text{GEO}, {\delta}^{\dagger}$ as described in Section \ref{sec:on-target}.
    \State \textbf{Step 3:} Solve $\widehat{\boldsymbol{\gamma}}$ according to \eqref{equ:emp:maximin} with $\delta=\delta^\dagger$.
    \State \textbf{Output}: $ \widehat{\tau}(\mathbf{x}) = \sum_{k=1}^K \widehat{\gamma}_k \widehat{\tau}_k(\mathbf{x})$.
\end{algorithmic}}
\end{algorithm}

The DRIFT estimator also yields an individualized treatment rule (ITR), $\widehat\pi(x) := I\{\widehat{\tau}(x) > 0\}$, which assigns treatment whenever our distributionally robust predictor for the effect is positive. As shown in Proposition \ref{propo:ITR}, this decision rule corresponds to the optimal regime under the maximin criterion for centered Gaussian covariates $\X$ and linear $\f^{(a)}, a = 0,1$.

{
\begin{Proposition}
\label{propo:ITR}
Suppose $\X$ follows a centered Gaussian distribution, then we have
\[
\tau^*\in\argmax{\tau\in\mathcal{F}}\min_{\boldsymbol{\gamma}\in\Gamma(\delta)} \mathbb{E}\left[\boldsymbol{\gamma}^\top(\Z^{(1)}-\Z^{(0)})\cdot I\left({\tau}(\X)>0\right)\right],
\]
where $\mathcal{F}$ is the linear function class. That is, the decision rule induced by the DRIFT estimator ${(i.e., I(\tau^*(\mathbf{x}) > 0))}$ corresponds to the optimal Individualized Treatment Rule (ITR) that maximizes the worst-case expected reward over the on-target (uncertainty) set.
\end{Proposition}

The optimal ITR characterized in Proposition \ref{propo:ITR} shares underlying technical similarities with the Maximin Projection Learning (MPL) framework proposed by  \cite{shi2018maximin}, particularly in utilizing a Gaussian design to establish a theoretical equivalence that facilitates the study of the estimator.
} 

\subsection{Observational Studies}\label{sec:method:obs}

In real-world applications, treatment assignments are often non-randomized. To properly adjust for confounding and estimate treatment effects in such observational settings, one could employ a two-stage doubly robust procedure (DR-Learner) adapted from \citep{kennedy2023doubleML}. After obtaining latent estimators $\{\widehat{\z}_i\}_{i=1}^N$ in Section~\ref{sec:fct_model}, we randomly divided the set of $N$ observations $\mathcal{D} := \{\x_i,a_i,\widehat{\z}_i\}_{i=1}^N$ evenly into two disjoint sets $\mathcal{D}_1,\mathcal{D}_2$ with an equal size $[N/2]$. Then we apply the DR-Learner method as follows:

\noindent\textbf{Stage 1: Nuisance training:} Construct estimates of two nuisance functions using $\mathcal{D}_1$: (i) $\widehat{s}$ of the propensity scores $s$, where $s(\x) = \mathbb{P}(A=1 \mid \X=\x)$; (ii) $(\widehat{\f}^{(0)},\widehat{\f}^{(1)})$ of the latent outcome regression functions $(\f^{(0)},\f^{(1)})$.

\vspace{0.15cm}

\noindent\textbf{Stage 2: Pseudo-outcome regression:} Construct the latent pseudo-outcome:
\begin{equation}
    \widehat{\varphi}(\Z)= \frac{A-\widehat{s}(\X)}{\widehat{s}(\X)[1-\widehat{s}(\X)]} \big\{\Z-\f^{(A)}(\X)\big\}+\widehat\f^{(1)}-\widehat\f^{(0)}
    \label{eq:dr_pseudo_outcome}
\end{equation}
and regress it on covariates $\X$ in the holdout sample $\mathcal{D}_2$, yielding $\widehat{\boldsymbol{\tau}}_{dr}(\x) = \widehat{\EE}[\widehat{\varphi}(\Z)\mid \X=\x]$.  The $k$-th element $\widehat{{\tau}}_{dr,k}(\x)$ is then used as the factor-wise treatment effect estimation, which serves as the input of (\ref{equ:emp:maximin}) for computing the final maximin estimator.

\section{Theoretical Properties}\label{sec:theory}

Let $\|\mathbf{M}\|_F = \big\{\sum_{i=1}^N \sum_{j=1}^JM_{ij}^2\big\}^{1/2}$ be the Frobenius norm of a matrix $\mathbf{M} \in\R^{N \times J}$ and denote the smallest and largest eigenvalues of a squared matrix $\bXi$ as $\lambda_{\min}(\bXi), \lambda_{\max}(\bXi)$, respectively. 
We use $\|\boldsymbol{u}\|_2$ to denote the $\ell_2$-norm of vector $\boldsymbol{u}$. For positive sequences $a(n), b(n)$, use $a(n)\lesssim b(n)$ when there exists some universal constant $c\geq 0$ such that $a(n)\leq cb(n)$ for all $n\geq 1$, and $a(n)\lesssim_{\mathrm{P}}b(n)$ for $a(n)\lesssim b(n)$ with the probability approaching one.

\subsection{Identifiability}
Before proceeding, we address the identification issue caused by the latent factor $\mathbf{Z}$ in the generalized factor model \eqref{eq:factor}. Note that model \eqref{eq:factor} is invariant under any affine transformations of $\mathbf{Z}$. Consequently, the unobserved $\Z$ is identifiable only up to an \textit{admissible reparameterization} $\tilde{\mathbf{Z}} = \mathbf{Q}\mathbf{Z} + \boldsymbol{\mu}$, where $\mathbf{Q} \in \R^{K \times K}$ is invertible and $\boldsymbol{\mu} \in \R^K$.  Despite the non-identifiability of $\Z$, the core components of our framework remain intrinsic to the data and invariant to the choice of latent representation. Specifically, it is straightforward to verify that for any admissible reparameterization $\tilde{\mathbf{Z}}$, the induced optimal factorized outcome for any given response $Y$, including GEO, is identical to $\phi_{Y}(\mathbf{Z})$ obtained with the original $\mathbf{Z}$. This invariance of the optimal factorized outcomes guarantees that the proposed maximin ITE estimator is well-defined, as formally stated in Proposition~\ref{propo:well_defined}.

\begin{Proposition}[Invariance of DRIFT to Latent Reparameterization]
\label{propo:well_defined}
    Let $\tilde{\mathbf{Z}} = \mathbf{Q}\mathbf{Z} + \boldsymbol{\mu}$ be any admissible reparameterization of $\mathbf{Z}$. For any $\phi\in\mH$, any response $Y$ and any radius $\delta\ge0$, we define the induced quantities as $ \tilde R_\phi, \tilde{\phi}_Y, \tilde{d}_Y$ and $\tilde{\Phi}(\delta)$ by replacing $\Z$ with $\tilde\Z$ in their respective definitions. The resulting DRIFT function $\tilde\tau^* :=\argmaxdum_{\tau\in\FF} \ \min_{\bgamma\in\tilde\Gamma(\delta)}\tilde R_\phi(\tau)$ is identical to the original $\tau^*$ in \eqref{eq:DRIFT} with the same radius $\delta$. Notably, setting $\delta=0$ establishes the invariance of the ITE for the GEO. 
\end{Proposition}

\subsection{Estimation Consistency}

We now establish the convergence of the ITE estimator proposed in
Section~\ref{sec:method_2} under an asymptotic regime in which both the
number of subjects $N$ and the number of items $J$ diverge ($N, J \to \infty$),
while the latent dimension $K$ remains fixed. We further assume that the
treatment and control sample sizes $N_1$ and $N_0$ satisfy
$\min\{N_1, N_0\}\asymp N$. Let $\mathbf{U}^* = [\mathbf{z}_1 \cdots \mathbf{z}_N]^\top \in \R^{N \times K}$ denote the matrix of true latent factors, and $\mathbf{W}^* = [\boldsymbol{\alpha}_1 \cdots \boldsymbol{\alpha}_J]^\top \in \R^{J \times K}$ represent the loading matrix, with the intercept vector $\boldsymbol{\zeta}^* = (\zeta_1, \dots, \zeta_J) \in \R^{J}$. Denote $(\widehat{\mathbf{U}}, \widehat{\mathbf{W}}, \widehat{\boldsymbol{\zeta}})$ as the corresponding solution to the empirical problem~\eqref{eq:JML}. 

Suppose the treatment effect model \eqref{eq:Z} admits a linear form: $\f^{(a)}(\X) = \bLambda^{(a)}\X, \ \bLambda^{(a)}\in\R^{K\times p}$ for $a = 0,1$. Let $\mathbf{B}^* = [\bbeta_1 \cdots \bbeta_K]:= (\bLambda^{(1)}-\bLambda^{(0)})^\top\in \R^{p \times K}$ denote the matrix of the true factor-wise linear effect parameters. Then, the factor-wise ITE function is given by $\tau_k(\x)= f_k^{(1)}(\x)-f_k^{(0)}(\x)=\bbeta_k^\top \x$, for $k\in[K]$. We let $\widehat{\mathbf{B}} = [\widehat{\bbeta}_1 \cdots \widehat{\bbeta}_K]\in \R^{p \times K}$ denote the estimator of $\B^*$ obtained by linearly regressing the recovered latent outcomes $\widehat{\U}$ on the covariates. Although this analysis primarily addresses the linear case, extensions to non-linear ITE models are feasible. We rely on some regularity and non-singularity assumptions to guarantee the convergence of  $\widehat{\mathbf{U}}$, $\widehat{\mathbf{W}}$ and $\widehat{\mathbf{B}}$, up to some invertible rotation and scaling. 

\begin{assumptionA}
    \label{ass:A1}
    The generalized factor model is correctly specified in problem~\eqref{eq:JML}, i.e., the true model parameters and latent factors satisfy that 
    $\sqrt{1+\|\mathbf{z}_i\|_2^2} \le C, \  \forall i \in [N]$ and $ 
    \sqrt{\zeta_j^2+\|\boldsymbol{\alpha}_j\|_2^2} \le C, \ \forall j \in [J],$ where $C$ is the boundary of the feasible set used in \eqref{eq:JML}.
\end{assumptionA}

\begin{assumptionA}\label{ass:A2}
    There exist constants $L,B> 0$ such that for all $j \in [J]$ and $y$ in the domain of $Y_j$, (i) Strong convexity and smoothness:
        $B^{-1} \le {\partial^2 \ell_{Y_j}(y,\phi)}/{\partial \phi ^2} \le B,$ $\forall \phi \in [-C^2, C^2]$; (ii) Lipschitz continuity: $|{\partial \ell_{Y_j}(y,\phi)}/{\partial \phi }|\le L, \forall \phi\in [-C^2, C^2]$. 
\end{assumptionA}

\begin{assumptionA}\label{ass:A3}
    For any response $Y \in \{Y_1, \cdots, Y_J, O\}$ and $\phi \in [-C^2, C^2]$, the centered loss process $\mathcal{L}_Y(\phi) := \ell_{Y}(Y, \phi) - \mathbb{E}[\ell_{Y}(Y, \phi)]$ satisfies either (i) or (ii): 
    \begin{enumerate}
        \item[(i)] Boundedness: $|\mathcal{L}_Y(\phi)| \le B'$ for some constant $B'> 0$;
        \item[(ii)] Sub-exponential: There exist $\nu, \alpha > 0$ such that $\mathbb{E}[e^{\lambda \mathcal{L}_Y(\phi)}] \le e^{\lambda^2\nu^2/2}$ for all $|\lambda| < 1/\alpha$.
    \end{enumerate}
\end{assumptionA}

\begin{assumptionA}\label{ass:A4}
    There exist some constant $\sigma> 0$ such that for $a\in\{0,1\}$,     $\lambda_{\min}\big(\Z\supa(\Z\supa)^{\top}\big) \geq \sigma^{-1}$, $\lambda_{\min}\big(J^{-1}\mathbf{W}^{*\top}\mathbf{W}^*\big) \ge \sigma^{-1}$, $\sigma^{-1}\le \lambda_{\min}(\bSigma) \le  \lambda_{\max}(\bSigma)\le \sigma$, $\lambda_{\max}\big(\EE[\boldsymbol\epsilon\supa (\boldsymbol\epsilon\supa)^\top]\big)\leq \sigma$, where $\bSigma := \EE[\mathbf X \mathbf X^\top]$ denotes the second-moment of the covariates $\mathbf X$ and $\boldsymbol\epsilon\supa$ is the noise term in model~\eqref{eq:Z}. 
\end{assumptionA}

Assumption \ref{ass:A1} guarantees that the estimator from the constrained optimization problem \eqref{eq:JML} is well-specified. Assumption \ref{ass:A2} imposes regularity conditions on the ordinal loss, ensuring it is well-behaved for optimization. Meanwhile, Assumption \ref{ass:A3} controls the noise level in the observed responses. Collectively, these three assumptions ensure that the estimated factorized outcome matrix, $\widehat{\mathbf{M}} = \widehat{\mathbf{U}}\widehat{\mathbf{W}}^{\top} - \mathbf{1}_N\widehat{\boldsymbol{\zeta}}^\top$, is consistent with the true matrix, $\mathbf{M}^* = \mathbf{U}^*\mathbf{W}^{*\top} - \mathbf{1}_N\boldsymbol{\zeta}^{*\top}$, in terms of the Frobenius norm. Assumption \ref{ass:A4} includes a non-degeneracy condition requiring strictly positive lower bounds on the eigenvalues of the latent covariance and weight matrices. This prevents the latent factor model from collapsing and guarantees that each loading weight retains sufficient information as the dimension $J$ of the items grows. Also, Assumption 4 requires non-singular covariates $\X$ and bounded noise variance, both of which are standard conditions for linear regression.

Based on these assumptions, Lemma~\ref{lemma:U_W_B} establishes that $\widehat{\mathbf{U}}$, $\widehat{\mathbf{W}}$ and $\widehat{\B}$ consistently recover the true latent factors, the weighting matrix of the generalized factor model (\ref{eq:factor}), and the factor-wise linear model coefficients. This recovery holds in a standard sense, up to an invertible scaled rotation. 
\begin{Lemma}\label{lemma:U_W_B}
   Let $\widehat{\mathbf{U}}_c = (\mathbf{I}_N - \mathbf{P}_{\mathbf{1}})\widehat{\mathbf{U}}$ and $\mathbf{U}^*_c = (\mathbf{I}_N - \mathbf{P}_{\mathbf{1}})\mathbf{U}^*$ denote the centered latent factor matrices, where $\mathbf{P}_{\mathbf{1}} = \frac{1}{N}\mathbf{1}_N\mathbf{1}_N^\top$.
    Under Assumptions \ref{ass:A1}-\ref{ass:A4}, when $N, J \rightarrow \infty$, we have
    \begin{align}
        \min_{\mathbf{Q} \in \mathcal{O}_K} \frac{1}{J} \| \widehat{\mathbf{W}}\mathbf{Q} - \mathbf{W}^* \|_F^2 +
        \min_{\mathbf{Q} \in \mathcal{O}_K} \frac{1}{N} \| \widehat{\mathbf{U}}_c\mathbf{Q} - \mathbf{U}_c^* \|_F^2+\min_{\Q\in\mO_K}\|\widehat{\B}\Q-\B^*\|_F^2 \lesssim_{\mathrm{P}} \sqrt{\frac{N+J}{NJ}}, \notag
    \end{align}
   where $\mathcal{O}_K = \{ \Q\mid\mathbf{Q} \in \R^{K \times K}, \det\Q \neq0 \}$.
\end{Lemma}

\begin{assumptionA}\label{ass:A5}
It is satisfied that $\lambda_{\min}(\mathbf{\Xi}) > 0$ where ${\Xi}_{k,l} := \EE[\tau_k(\X)\tau_l(\X)]$ for $k,l \in[K].$
\end{assumptionA}

While $\mathbf{\Xi}$ defined in~\eqref{eq:quad} is inherently positive semi-definite, Assumption~\ref{ass:A5} requires strict positive definiteness to control the uncertainty introduced by estimating $\bXi$, and hence bound the error of the solution to our maximin problem. Ridge regularization could be potentially used to enforce the well-posedness of $\bXi$ in the presence of the singularity \citep{guo2024statistical}. In Theorem \ref{thm:convergence}, we establish the convergence of the proposed DRIFT estimator.
\begin{Theorem}\label{thm:convergence}
    Suppose Assumptions \ref{ass:A1}-\ref{ass:A5} hold. For a given radius $\delta>0$, let $\tau^*(\x) = \bbeta^{*\top}\x $ be the true maximin ITE over the uncertainty set $\Gamma(\delta)$, and $\widehat{\tau}(\x) = \widehat{\bbeta}^{\top}\x$ be the estimator constructed using $\widehat{\mathbf{U}}$ over the uncertainty set $\widehat\Gamma(\delta)$. Then we have:
    $$\|\widehat\bbeta-\bbeta^*\|_2\lesssim_\rmp \sqrt{\tilde{E}}\cdot \max\Big\{\sqrt{\tilde{E}},\frac{1}{\sqrt{\delta}}\Big\},  \text{as } N,J \to \infty,$$
where $\tilde{E}=(\frac{N+J}{NJ})^{1/4}+(\frac{1}{N})^{1/2}.$
\end{Theorem}

Theorem \ref{thm:convergence} shows that the convergence rate of our empirical estimator $\widehat{\tau}(\x)$ can be decomposed into two components: the first term captures the estimation error of the linear coefficient matrix $\B^*$. {The second term accounts for the additional uncertainty in estimating the optimal aggregation weights $\boldsymbol{\gamma}$, arising from the empirical approximation of $\bXi$ and the on-target set $\Gamma(\delta)$.} 
This result distinguishes our theory from previous work on maximin regression (e.g., \cite{meinshausen2015maximin,guo2024statistical}) by resolving the challenge posed by estimating the on-target set $\Gamma(\delta)$. Our explicit control of the estimation error for this set is essential for the consistency of the maximin ITE estimator. Moreover, this theoretical guarantee is not limited to linear parametric models. When more flexible machine learning methods are used to fit factorized outcomes in high-dimensional settings, similar convergence results can also be obtained \citep{wang2023distributionally}.

\section{Simulation Study}
\label{sec:simul}

\subsection{Experiment Setup}

We conduct numerical experiments to evaluate the finite-sample performance of the proposed estimator. For each subject $i\in[N]$ we generate the baseline covariate vector $\mathbf{x}_{i}\in\R^5$, where ${\x}_{i,1:4}\sim \mathcal{N}(\mathbf{0}, \I_4)$ and $\mathrm{x}_{i,5}\equiv 1$ represents the intercept term. The randomized treatment assignment follows a Bernoulli design with $P(a_i=1)=0.5$. The potential latent factors ($K = 3$) follow a linear model: $\mathbf{z}_i^{(a)} = \boldsymbol{\Lambda}^{(a)}\mathbf{x}_i + \boldsymbol{\epsilon}^{(a)}$, where $\boldsymbol{\epsilon}^{(a)} \sim \mathcal{N}(\mathbf{0}, \mathbf{I}_K)$ and $\boldsymbol{\Lambda}^{(a)}$ are fixed matrices. The realized factor is $\mathbf{z}_i = a_i\mathbf{z}_i^{(1)} + (1 - a_i)\mathbf{z}_i^{(0)}$.

The binary GEO is generated as $o_i \sim \text{Bernoulli}(\text{expit}(\boldsymbol{\gamma}_{\text{GEO}}^\top \mathbf{z}_i))$ with $\boldsymbol{\gamma}_{\text{GEO}} = (1, 1, 1)^\top$. We simulate $J = 30$ binary responses $Y_{ij}$ using Eq~(\ref{eq:factor}). To introduce controlled heterogeneity, the outcome loadings $\boldsymbol{\alpha}_j$'s are drawn from a hyper-population centered around $\boldsymbol{\gamma}_{\text{GEO}}$:
\begin{equation}
    \label{set:alpha}
    \boldsymbol{\alpha}_j
    =\boldsymbol{\gamma}_{\mathrm{GEO}}
    + r\|\boldsymbol{\gamma}_{\mathrm{GEO}}\|_2\, R_j\,\boldsymbol{v}_j,
    \quad
    R_j\sim \mathrm{Beta}(5,1.5),
    \quad
    \|\boldsymbol{v}_j\|_2=1.
\end{equation}
Here, $\boldsymbol{v}_j$ follows a von Mises-Fisher (vMF) distribution on the unit sphere with concentration parameter $\sigma_v > 0$ around the mean direction $\boldsymbol{\gamma}_{\text{GEO}}/\|\boldsymbol{\gamma}_{\text{GEO}}\|_2$. The scalar $r$ bounds the maximum relative $\ell_2$ deviation of $\boldsymbol{\alpha}_j$ from $\boldsymbol{\gamma}_{\text{GEO}}$. Together, $r$ and $\sigma_v$ explicitly govern the dispersion of the loading vectors, controlling the similarity between the $J$ outcomes and GEO.

In the following analysis, we fix the sample sizes $N=300$ and evaluate the robustness of \textit{DRIFT} to settings with different degrees of (i) variance within the hyper-population and (ii) observed-outcome underrepresentation. The effect of sample size on estimator stability is studied separately in Appendix. In each setting, we estimate the ITE using the observed data $\{(\mathbf{x}_i,\mathbf{y}_i,a_i,o_i)\}_{i=1}^N$ and compare the following methods: (1) \textit{Original GEO}: ITE for the GEO is learned through an R-learner \citep{r_learner} on the log-odds scale; (2)  \textit{Factorized GEO}: ITE is learned through the combination of latent factors guided by the GEO; (3) \textit{DRIFT}: maximin ITE estimator from the proposed framework by applying
Algorithm~\ref{algo:maximin} with $\{o_i\}_{i=1}^N$ observed; (4)  \textit{DRIFT\_N}: same as \textit{DRIFT} with $\{o_i\}_{i=1}^N$ unobserved. (5) \textit{Obs Maximin}: (only for observed-outcome underrepresentation setting) ``observed-only'' maximin ITE estimator over the convex hull of observed factorized outcomes. 

To assess generalizability, we evaluate each method on $T=1,000$ external outcomes $\{\mathring{Y}_t\}_{t=1}^T$ independently sampled from the hyper-population. For each $t$, $\mathring{Y}_t$ follows the generalized factor model in \eqref{eq:factor}, with loading vector $\mathring{\boldsymbol{\alpha}}_t$ drawn i.i.d.\ from the same distribution as the in-sample item loadings $\{\boldsymbol{\alpha}_j\}_{j=1}^J$. For a subject with covariates $\mathbf{x}$, the true ITE for $\mathring{Y}_t$ is \(
\mathring{\tau}_t(\mathbf{x})
= \mathring{\boldsymbol{\alpha}}_t^{\top}\big(\boldsymbol{\Lambda}^{(1)}-\boldsymbol{\Lambda}^{(0)}\big)\mathbf{x}.
\)
We summarize performance using the worst-case prediction accuracy 
$\mathrm{ACC}_{\min} := \min_{ t\in [T]}\mathrm{ACC}_t$, and correlation $\mathrm{Cor}_{\min} := \min_{ t\in [T]}\mathrm{Cor}_t$ between the predicted and true ITEs across external outcomes, where 
$\mathrm{ACC}_t
=\frac{1}{N}\sum_{i=1}^N
I\!\left\{\operatorname{sign}\!\big(\widehat{\tau}(\mathbf{x}_i)\big)
=\operatorname{sign}\!\big(\mathring{\tau}_t(\mathbf{x}_i)\big)\right\},$ and $\mathrm{Cor}_t$ is the empirical correlation between $\widehat{\tau}(\mathbf{x}_i)$ and $\mathring{\tau}_t(\mathbf{x}_i)$.
For each setting, we repeat the simulation
100 times and report the worst-case metrics ($\mathrm{ACC}_{\min}$, $\mathrm{Cor}_{\min}$) for each method.

\subsection{Robustness to Variation of the Outcome Representation}

\begin{figure}[ht]
     \centering
    \includegraphics[width=0.9\textwidth]{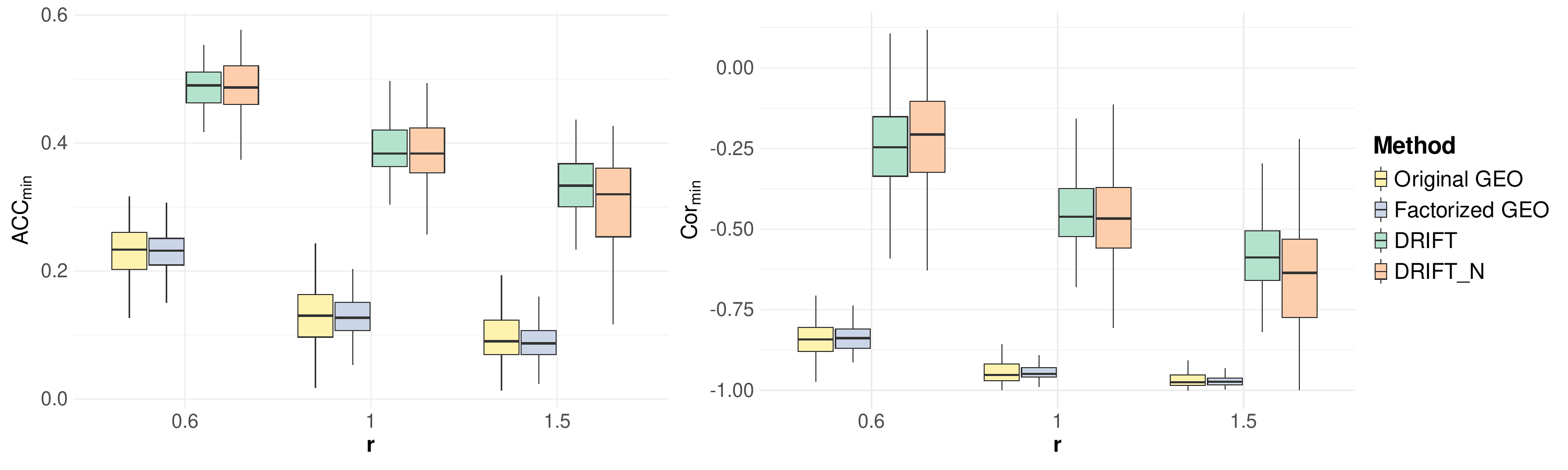}
      \caption{Boxplots of $\mathrm{ACC}_{\min}$ (left) and $\mathrm{Cor}_{\min}$ (right) for the predicted individual treatment effects (ITEs) over 100 replications. In each replication, $J=30$ observed responses and $T=1000$ external responses are sampled from the loading hyper-population with fixed concentration $\sigma_v=1$, while the deviation magnitude varies as $r\in\{0.6,1,1.5\}$. The subject sample size is fixed at $N=300$.}
     \label{fig:simulr}
\end{figure}

To mimic different degrees of variance within the hyper-population of the outcome representations \eqref{set:alpha}, we fix the concentration at \(\sigma_v=1\) while varying deviation magnitude \(r\in\{0.6,1,1.5\}\). Each time, we draw the loading vectors for observed outcomes (\(\boldsymbol{\alpha}_j\)'s) and the external outcomes (\(\mathring{\boldsymbol{\alpha}}_t\)'s) 
i.i.d.\ from the hyper-population with the same $r$. We repeat the simulation 100 times for each $r$, and report the worst-case performance metrics in Figure~\ref{fig:simulr}. The results show that the worst-case ACC and Cor decrease across all methods as $r$ increases. This trend is expected, since a larger $r$ allows for greater divergence among symptom outcomes, introducing additional uncertainty into ITE prediction. Across $r \in \{0.6,1,1.5\}$, \textit{DRIFT} achieves consistently higher \(\mathrm{ACC}_{\min}\) and substantially less negative \(\mathrm{Cor}_{\min}\) than \textit{Original GEO} and \textit{Factorized GEO} in average, indicating superior ability to exploit the hyper-population structure for worst-case optimization even with increased outcome heterogeneity. { For example, when $r=1$, \textit{DRIFT} achieves a mean ACC$_{\min}$ of $0.420$, outperforming \textit{Original GEO} and \textit{Factorized GEO} by $154.5\%$ and $156.1\%$, respectively. Meanwhile, in terms of mean Cor$_{\min}$, \textit{DRIFT} exceeds \textit{Original GEO} 
and \textit{Factorized GEO}
by $58.3\%$ and $58.1\%$.} 

Notably, \textit{DRIFT\_N} performs comparably to \textit{DRIFT} in each setting, suggesting that our proposed framework remains effective even without observed GEO and is practically applicable. However, \textit{DRIFT\_N} shows larger variance in both metrics than \textit{DRIFT}, reflecting the added uncertainty from estimating GEO to construct the on-target set. In Appendix, we also numerically justify our choice $\delta=\delta^\dagger$ for $\Gamma(\delta)$. Evaluating DRIFT across $\delta \in \{0.2\delta^\dagger, 0.5\delta^\dagger, \delta^\dagger, 1.5\delta^\dagger\}$, the worst-case performance ($\text{ACC}_{\min}$ and $\text{Cor}_{\min}$) peaks at $\delta^\dagger$ and decreases for larger value. This suggests $\delta^\dagger$ optimally balances capturing relevant external information and excluding uninformative ones that could reduce GEO's guidance.

\subsection{Robustness to Underrepresentation in Observed Outcomes}
As discussed in Section~\ref{sec:method:geo}, \textit{Obs Maximin} can be unstable when the observed outcomes provide only partial coverage of the ``hyper-population'', especially when the loading vectors \(\boldsymbol{\alpha}_j\) are concentrated in a restricted region (Figure~\ref{fig1:maxmin}). In contrast, \textit{DRIFT} is designed to adapt to such under-coverage. To illustrate this advantage, we generate the observed-outcome loadings \(\boldsymbol{\alpha}_j\) from \eqref{set:alpha} with $r=0.6$ and larger concentration parameters
\(\sigma_v\in\{2,4,8\}\), which restricts \(\boldsymbol{\alpha}_j\) to a narrower cone around the mean direction. The external outcomes used for evaluation are sampled from the hyper-population with scalar $r=0.6$ and concentration fixed at \(\sigma_v=1\). We compare our proposed method (\textit{DRIFT} and \textit{DRIFT\_N}) with \textit{Original GEO}, \textit{Obs Maximin}, and \textit{Factorized GEO} using \(\mathrm{ACC}_{\min}\) and \(\mathrm{Cor}_{\min}\) over 100 replications.
\begin{figure}[ht]
     \centering
    \includegraphics[width=0.9\textwidth]{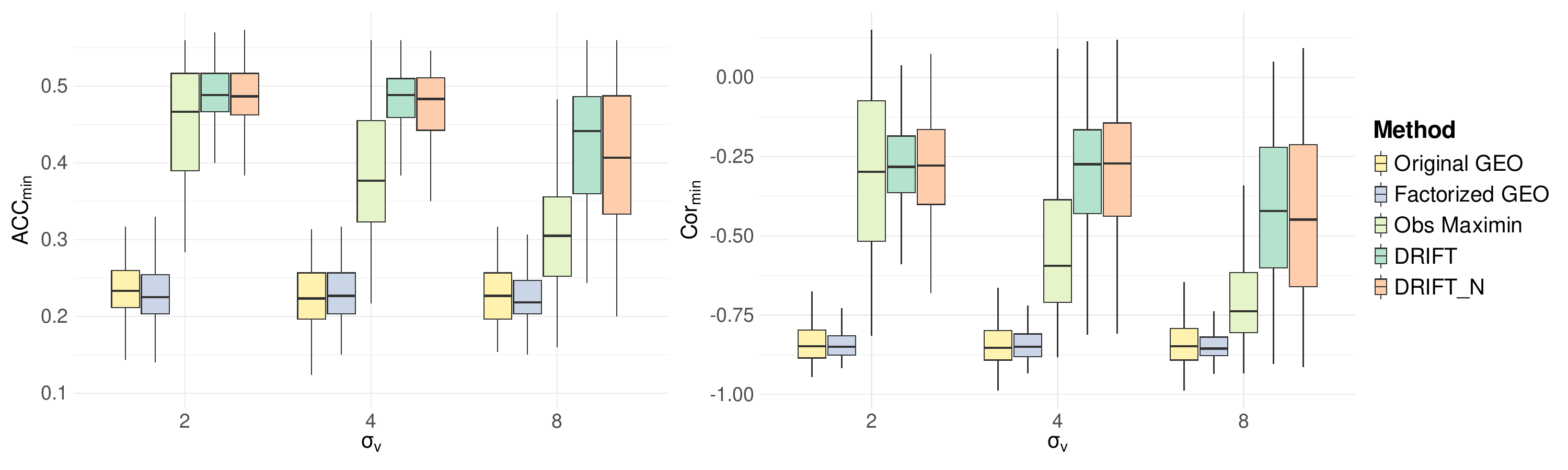}
     \caption{Boxplots of $\mathrm{ACC}_{\min}$ (left) and $\mathrm{Cor}_{\min}$ (right) for predicted ITEs over 100 replications. We fix sample size $N=300$ and scalar $r=0.6$. The $J=30$ observed-response loadings are generated with concentration at \(\sigma_v\in\{2,4,8\}\), while $T=1000$ external responses are sampled from the hyper-population with \(\sigma_v=1\).} 
     \label{fig:kappa_dif}
\end{figure}

Figure~\ref{fig:kappa_dif} shows that \textit{Original GEO} and \textit{Factorized GEO} are essentially insensitive to \(\sigma_v\), but their worst-case performance remains poor because they do not explicitly account for
on-target outcome uncertainty. \textit{Obs Maximin} attains comparatively strong \(\mathrm{ACC}_{\min}\) and \(\mathrm{Cor}_{\min}\) when \(\sigma_v\) is small, deteriorates sharply as \(\sigma_v\) increases, indicating its sensitivity to mismatch between the observed outcomes and the evaluation hyper-population. In contrast, \textit{DRIFT} and \textit{DRIFT\_N} exhibit substantially slower deterioration with increasing \(\sigma_v\) and achieve the best worst-case performance overall, demonstrating robustness when the observed outcomes are underrepresenting the hyper-population.

\section{Application to EMBARC}\label{sec:real}

\subsection{Background of the Study}
We applied \textit{DRIFT} to a randomized controlled trial of major depressive disorder (MDD), EMBARC \citep{trivedi2016establishing}, to assess its real-world performance. 
Participants were randomly assigned to receive either sertraline ($A=1$) or placebo ($A=0$).
The Clinical Global Improvement (CGI) scale assessed at the end of treatment was used as the GEO to guide the construction of the on-target factorized outcome set. 
Individual items in the Hamilton Depression Rating Scale (HAM-D) \citep{hamilton1960rating}, a clinician-administered depression assessment, were used as the manifest outcomes $\mathbf{Y}$ to learn latent factors $\mathbf{Z}$.
To enhance learning efficiency and maintain sample balance, items with few endorsements and inconsistent collection procedures were excluded  \citep{van2015association, chekroud2017reevaluating}, and sparse categories were combined. 
Electroencephalogram (EEG) measures of brain activity related to antidepressant response were also incorporated to improve the estimation performance on the latent factors \citep{yang2024learning}. 
{Because the excess loss for constructing the on-target set is defined for binary GEO, including representations of continuous EEG outcomes would lead to scale incompatibility. Therefore, we construct the on-target set using only HAM-D items and use EEG measures only to improve latent factor estimation.}
Given that prior studies identified 2–4 latent factors for depression  \citep{van2016latent, van2012data, sunderland2013factor,chen2021learning}, we fit the factor model with 4 latent domains.
Covariates $\mathbf{X}$ include age, sex, Flanker Interference Accuracy score (indicating cognitive control), and inferred baseline latent domains. 
After excluding participants with missing measures, 168 subjects were included in the analysis, of whom 75 received sertraline.

We compared our method with three alternatives: \textit{Original GEO}, \textit{Factorized GEO} and \textit{Obs Maximin} (see Section~\ref{sec:simul}). Because \textit{Obs Maximin} suffers from latent factor non-identifiability, we applied an oblique promax rotation \citep{promax} to the loading matrix and used the rotated structure as the factor estimates for fair comparison. {It also improves clinical interpretability by allowing each rotated factor to better reflect distinct, clinically meaningful domains.} To assess generalizability, we assessed the prediction accuracy (ACC in Section~\ref{sec:simul}) of each method on external outcomes not used in training. These external outcomes include (i) the self-reported 17-item Concise Associated Symptom Tracking (CAST) scale, a validated instrument for quantifying depression-related symptoms by probing experiences such as anxiety, sleep disturbances, restlessness, and shifts in mood and energy \citep{minhajuddin2020psychometric}, and (ii) the self-rated Frequency, Intensity, and Burden of Side Effects Rating (FIBSER) scale, which measures medication side effects \citep{wisniewski2006self}. Because these outcomes differ clinically from the training responses, performance on them reflects out-of-distribution generalizability. 

\subsection{Evaluation of ITE Estimation}

Table~\ref{tab:real_res} summarizes selected results of prediction accuracy on external outcomes for all four methods. For 17 CAST items, we compute ACC for each item and report four lowest-ranked items (ranked by ACC). For FIBSER self-rated side effects, we report ACC for every item (Frequency, Intensity, and Interference). The findings can be summarized as two key points:

\textbf{Protection against domain-specific failures:} Relying on a single outcome limits both the \textit{Original} and \textit{Factorized GEO}, causing concentrated failure clusters in the mania domain (CAST $7, 2, 17, 8$, ranked from lowest to fourth-lowest in ACC). Using the maximin strategy, \textit{Obs Maximin} partially mitigates these domain-specific issues by improving ACC on the two worst items (CAST $7,2$) and shifting the fourth lowest-ranked item to the anxiety domain (CAST $6$). Nevertheless, by extrapolating to a broader range of symptoms, \textit{DRIFT} achieves a more balanced performance across symptom domains, effectively avoiding systematic domain-specific degradation. First, it improves ACC on the two most challenging mania items (CAST $2, 7$) by over $23.8\%$ compared to the other three alternatives (ACC of \textit{DRIFT}: $\ge0.343$, ACC of others: $\le 0.277$). Furthermore, the four lowest-ranked items of \textit{DRIFT} are no longer dominated by the mania domain, with the third and fourth lowest-ranked items shifting to the anxiety-related items CAST $6$ and $15$. 

{Robust prediction in the mania domain is particularly important when evaluating the treatment effect of sertraline for MDD. In this context, a substantial increase in mania-related scores may be a safety concern, rather than as part of the intended therapeutic benefit of sertraline. An ITE estimator that performs poorly in this domain risks failing to identify patients vulnerable to antidepressant-induced manic symptoms, leading to unsafe clinical recommendations. {By GEO-anchored extrapolation, \textit{DRIFT} prevents recommending treatments that alleviate depressive symptoms at the cost of triggering treatment-emergent mania, ensuring safer personalized decisions.}
}

\textbf{Generalization to side effects:} \textit{DRIFT} demonstrates stronger generalizability to side effect items in FIBSER (ACC: $0.307\text{--}0.319$), exceeding the two GEO-based baselines by more than $131\%$ (ACC: $\le0.133$) and the \textit{Obs Maximin} baseline by over $15.8\%$ (ACC: $0.217\text{--}0.265$). {This pattern suggests that \textit{DRIFT} is not only better at detecting the presence of adverse effects, but also more capable of identifying whether those adverse effects are frequent, severe, and disruptive enough to undermine adherence, all of which are key drivers of treatment discontinuation in clinical practice. {By anticipating side effects, \textit{DRIFT} addresses the narrow focus of standard MDD outcomes on core symptoms alone, and reduces the risk of recommending treatments that are effective but poorly tolerated.}

\begin{table}[t]
\centering
\footnotesize
\resizebox{\linewidth}{!}{%
\begin{tabular}{lcccc ccc}
\toprule
& \multicolumn{4}{c}{\makecell{\textbf{CAST} self-reported symptoms}}
& \multicolumn{3}{c}{\makecell{\textbf{FIBSER} self-rated side effects}} \\
\cmidrule(lr){2-5}\cmidrule(lr){6-8}
Method & 1st & 2nd & 3rd & 4th & Frequency & Intensity & Interference \\
\midrule
Original
& 0.066 & 0.108 & 0.367 & \textbf{0.482}
& 0.060 & 0.120 & 0.000 \\
Factorized GEO
& 0.102 & 0.145 & 0.392 & \textbf{0.494}
& 0.084 & 0.133 & 0.036 \\
Obs Maximin & 0.271 & 0.277 & \textbf{0.488} & \textbf{0.494} & 0.241 & 0.265 & 0.217\\
DRIFT
& \textbf{0.343} & \textbf{0.361} & \textbf{0.476} & \textbf{0.482} 
& \textbf{0.319} & \textbf{0.307} & \textbf{0.307} \\
\bottomrule
\end{tabular}}
\caption{Prediction accuracy (ACC) on external outcomes across items. For CAST, we list the four lowest-ranked items out of 17 ($1$st to $4$th): CAST ${7,2,17,8}$ for \textit{Original GEO} and \textit{Factorized GEO}, CAST ${7,2,17,6}$ for \textit{Obs Maximin} and CAST ${2,7,6,15}$ for \textit{DRIFT}. {Notably, CAST ${7,2,17,8}$ are mania-related, while CAST ${6,15}$ are anxiety-related items.} For FIBSER, we report the ACC for each of the three items (Frequency, Intensity, and Interference). {The best ACC and any comparable results within a 0.02 margin are highlighted in bold for each column.}
}
\label{tab:real_res}
\end{table}

In the subsequent section, we introduce the geometric visualization of the observed-outcome space and the solution of methods under comparison. This could offer deeper insight into the mechanisms driving the improvements of our approach over other methods.

\subsection{Geometric Visualization and Interpretation}
{To interpret the learned domains, we visualize the promax-rotated loading matrix with symptom-wise max-absolute scaling in Figure~\ref{fig:heatmaps}, which highlights each factor’s relative contribution to HAM-D symptoms. Factor 1 reflects a core depression domain, with strong loadings on depressed mood, guilt, loss of interest, and somatic anxiety. Sleep disturbances are split into two components: Factor 2 captures sleep-onset problems (initial insomnia), whereas Factor 3 captures sleep-maintenance difficulties (middle and late insomnia). Factor 4 represents a severe psychopathology and anxiety domain, marked by pronounced loadings on suicide, psychic anxiety, agitation, and retardation, consistent with a high-risk profile of acute distress. Together, these latent factors summarize diverse mental-health domains spanning psychological, somatic, and cognitive symptoms \citep{posner2005circumplex, van2016latent}.}

\begin{figure}[htb!]
  \centering
  \begin{subfigure}[t]{0.4\textwidth}
    \centering
    \includegraphics[width=\linewidth]{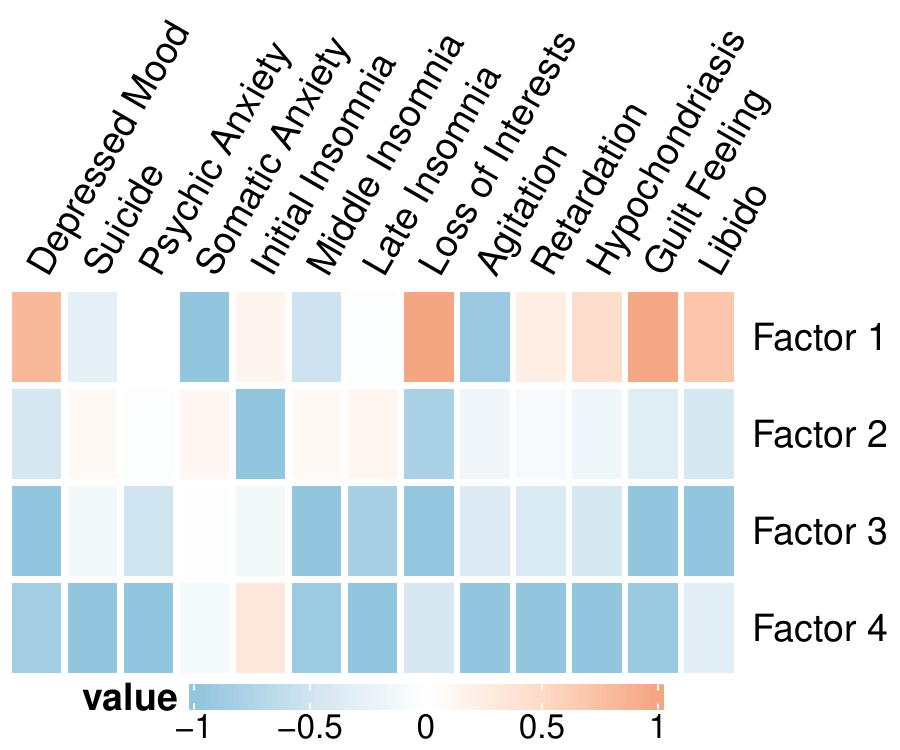}
    \caption{Heatmap of promax-rotated pattern loadings with symptom-wise max-absolute scaling. The value represents the relative contribution of each factor to individual clinical outcomes.}
    \label{fig:heatmaps}
  \end{subfigure} \hspace{1em}
  \begin{subfigure}[t]{0.52\textwidth}
     \centering
      \includegraphics[width=\textwidth]{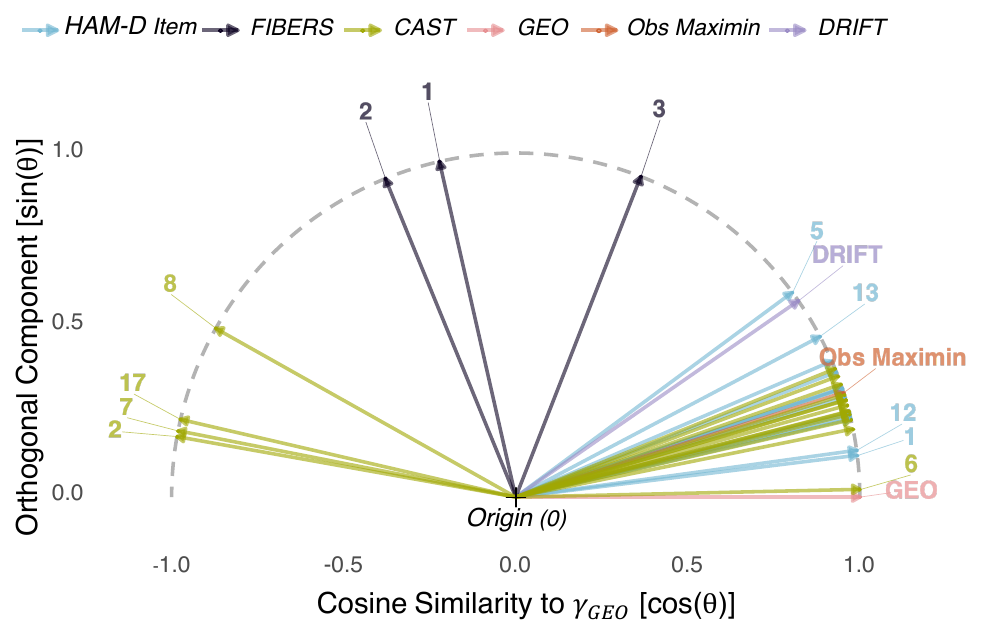}
         \caption{Cosine similarity (to $\bgamma_\GEO$) of estimated loading parameters $\balpha_j$'s for observed HAM-D items,  $\mathring{\balpha}_t$'s for external evaluation outcomes from FIBERS and CAST, the aggregation weight estimates $\bgamma_{\mathrm{obs}}$, $\bgamma_{\mathrm{DRIFT}}$.}
\label{fig:gamma_dis_item}
  \end{subfigure}
  \caption{Loading Analysis: Latent Domain Interpretation and Geometric Decomposition}
\end{figure}

{Furthermore, Figure~\ref{fig:gamma_dis_item} geometric summary of the estimated loading vectors for the observed HAM-D items and the external outcomes from FIBERS and CAST, together with the aggregation directions corresponding to \textit{Factorized GEO}, \textit{Obs Maximin}, and \textit{DRIFT}. The figure is shown in angular coordinates relative to the GEO direction: the x-axis gives the cosine similarity between each vector and $\gamma_{\text{GEO}}$, and the y-axis gives the corresponding orthogonal component through the sine value. 
Among the observed outcomes, those with large positive cosine value projections (e.g., depressed mood [HAM-D 1], guilt feeling [HAM-D 12]) are strongly aligned with the GEO direction, indicating that they largely move together with the GEO. In contrast, outcomes with significant orthogonal deviation (e.g., initial insomnia [HAM-D 5], libido [HAM-D 13]) capture complementary variation that is not well explained by the GEO alone. {The discrepancy is more pronounced for the external outcomes. As shown in Figure~\ref{fig:gamma_dis_item}, the side-effect outcomes (FIBERS 1-3) are substantially more orthogonal to GEO, whereas several mania-related outcomes (CAST 2, 7, 8, 17) lie on the negative-cosine side, indicating strong geometric misalignment with the GEO direction. This pattern suggests that the external outcomes contain clinically relevant variation that is largely underrepresented in the observed HAM-D subset, which in turn helps explain why ITE estimation based only on the GEO performs poorly for these domains.

This geometric interpretation highlights the advantage of \textit{DRIFT}. As shown in Figure~\ref{fig:gamma_dis_item}, whereas \textit{Obs Maximin} remains concentrated around the observed outcomes (blue arrows) and thus inherits their representational limitations, \textit{DRIFT} accounts for a broader range of symptom variation and better captures boundary information. Consequently, it yields an estimate with improved generalizability to underrepresented domains.} 
}

\section{Discussion}
\label{s:discuss}
The proposed DRIFT framework offers a principled approach to learning robust ITE across multiple outcome domains, with broad potential applications in precision medicine. Several extensions could further enhance its utility.
First, replacing the explained-variance reward function with an absolute-deviance criterion may reduce conservative estimates for small $\bgamma$ values and improve robustness to outliers. Second, the current linear treatment response model for latent factors could be relaxed to incorporate nonlinear CATE models such as spline regressions, or causal forests \citep{wager2018estimation}. Finally, addressing missing data in the items is an important extension. Multiple imputation or matrix completion \citep{davenport20141,chen2019joint} can be incorporated.

Another key extension is to directly derive individualized treatment rules (ITRs), using policy learning or value function optimization \citep{chen2021learning} to guide treatment strategies. Finally, expanding the framework to handle multiple treatments, including combination therapies or sequential, multiple assignment randomized trials (SMART) \citep{lei2012smart}, would offer the ability to estimate dynamic treatment regimens.

{
\bibliographystyle{apalike}
\bibliography{references} 
}

\end{document}